\documentclass[10pt,twocolumn,letterpaper]{article}

\usepackage{cvpr}
\usepackage{times}
\usepackage{epsfig}
\usepackage{graphicx}
\usepackage{amsmath}
\usepackage{amssymb}

\usepackage[pagebackref=true,breaklinks=true,letterpaper=true,colorlinks,bookmarks=false]{hyperref}

\cvprfinalcopy 

\def\cvprPaperID{****} 

\begin{document}

\title{Deep Image-to-Recipe Translation\\
}

\author{Jiangqin Ma, Bilal Mawji, Franz Williams\\
Georgia Institute of Technology\\
{\tt\small \{jma416, bmawji3, fwilliams70\}@gatech.edu}
}

\maketitle

\begin{abstract}
The modern saying, "You Are What You Eat," resonates on a profound level, reflecting the intricate connection between our identities and the food we consume. Our project, Deep Image-to-Recipe Translation, is an intersection of computer vision and natural language generation that aims to bridge the gap between cherished food memories and the art of culinary creation. Our primary objective involves predicting ingredients from a given food image. For this task, we first develop a custom convolutional network and then compare its performance to a model that leverages transfer learning. We pursue an additional goal of generating a comprehensive set of recipe steps from a list of ingredients. We frame this process as a sequence-to-sequence task and develop a recurrent neural network that utilizes pre-trained word embeddings.



We address several challenges of deep learning including imbalanced datasets, data cleaning, overfitting, and hyperparameter selection. Our approach emphasizes the importance of metrics such as Intersection over Union (IoU) and F1 score in scenarios where accuracy alone might be misleading. For our recipe prediction model, we employ perplexity, a commonly used and important metric for language models. We find that transfer learning via pre-trained ResNet-50 weights and GloVe embeddings provide an exceptional boost to model performance, especially when considering training resource constraints.

Although we have made progress on the image-to-recipe translation, there is opportunity for future exploration with advancements in model architectures, dataset scalability, and enhanced user interaction.

The code and datasets used in this study are available at: 
https://github.com/majiangqin/Image-to-Recipe-Translation-using-AI/tree/main.

\end{abstract}

\section{Introduction}

Just as the saying goes, "You Are What You Eat," the meaning behind this phrase is truer in more ways than one. While our bodies are composed of the nutrients we consume, our identities are also shaped by the ingredients, the preparation, and the cooking of food we make daily. Recipes that have been passed down generation to generation have meaning in our lives – they are the fond memories of family members who have shared meals with us. 


The goal of Deep Image-to-Recipe Translation is to give those who have pictures of their favorite dishes a chance to recreate a recipe for such a dish when they do not have the recipe in their possession. We seek to provide a set of ingredients and cooking instructions for a recipe that matches the food in a provided image. We would hope to give someone a chance to enjoy the same bite of food that their mother made for them years ago when they were a child.

This technique could be used by both novice cooks and professional chefs alike. One of the first things typically asked when experiencing an amazing dish is \emph{"How did you make it?"} --- the fascination and desire to recreate a dish from a fond memory is an all too common human experience. Development of an Image-to-Recipe Translation system would give people around the world the ability to recreate their cherished dishes and memories.

The task of generating ingredients and cooking instructions from an image is an intersection of two machine learning fields: computer vision (CV) and natural language generation (NLG). Deep CV methods have improved considerably since the advent of convolutional layers. Deep NLG has traditionally used recurrent layers, but more recently, attention-based methods have proven to be more powerful. Transfer learning also allows for reuse of previously trained networks to extract bottleneck features as part of a deep learning pipeline. We draw inspiration from and seek to reproduce aspects of the architecture found in \cite{Salvador_Drozdzal_Giro-i-Nieto_Romero_2019_orig_2}.

Our dataset comes from Kaggle \cite{Goel_Desai_Tanvi_2021_Kaggle} where the data was compiled by scraping the Epicurious website and specifically recipes with images attached to them. This dataset has been used in both academic and research settings and is also well-documented and of high quality. Each instance in the data consists of an ID number, a title, an ingredients list, an instructions list, an image name, and a cleaned ingredients list. There are in total around 13,500 instances within our dataset. Each instance's image name maps to an image file provided with the dataset. While this dataset was mostly clean, there were a few instances that contained null values which were removed as part of a data cleaning step. Additionally, extra preprocessing steps were also needed to clean the ingredients prior to filtering out stop words.

It's important to note that our dataset is more limited in scale (approximately 240 Megabytes) than other datasets such as Recipe1M (500 Gigabytes), which was used in \cite{Salvador_Drozdzal_Giro-i-Nieto_Romero_2019_orig_2}. The performance of deep learning models often benefits from extensive data. While our dataset constraints may contribute to the model's current limitations, it is a pragmatic compromise given hardware limitations.

Furthermore, the targets for our dataset were the list of ingredients and instructions. As previously mentioned, while reviewing the dataset, we came across the fact that our ingredients list needed to be refined to capture a better level of information.

Lastly, our work was inspired by researchers at Meta who have worked on a similar project. Meta researchers utilized similar models such as Convolutional Neural Networks (CNNs) and Recurrent Neural Networks (RNNs) for the type of data we use, but the implementations do differ. In the research done by Meta in \cite{Salvador_Drozdzal_Giro-i-Nieto_Romero_2019_orig_2}, they reference existing solutions for this task that rely on retrieval-based methods. In their words, "a recipe is retrieved from a fixed dataset based on the image similarity score in an embedding space". While this is a good method for predicting ingredients from an image, the researchers in \cite{Salvador_Drozdzal_Giro-i-Nieto_Romero_2019_orig_2} also mention that a good embedding system is required for retrieval to function correctly. One such limitation is that when an image cannot be found in the dataset, it can make retrieval for that recipe an impossible task.




\section{Approach}

Our approach to Image-to-Recipe Translation is composed of two stages. Stage 1 is our primary objective and involves predicting ingredients from a food image. Stage 2 involves generation of a full set of recipe steps and recipe title from the food image and predicted ingredients list. For this project, we utilized the TensorFlow deep learning library and its Keras high-level interface. Additionally, Scikit-learn, Pandas, Seaborn, and Matplotlib were used for data prepration, metrics, and visualization.

One problem we anticipated was the time required to train our ingredient prediction models. Both our custom CNN and ResNet-50 architectures involve convolutional layers, which are known to be computationally expensive. A second concern was our dataset size --- while many deep learning solutions rely on a large dataset, such a dataset would prevent us from training a model given our time constraints.

\subsection{Stage 1: Ingredient Prediction}

\subsubsection{Data collection and preprocessing}
For Stage 1, we augmented our data and preprocessed food images in an effort to improve our model's generalization. Our preprocessing involved resizing, normalizing, cropping, mirroring, rotating, flipping, and whitening images. Additionally, it came to our attention that we needed to refine each recipe's ingredients list. We preprocessed ingredients by merging those that either share two words at the start or end of their name as in \cite{Salvador_Drozdzal_Giro-i-Nieto_Romero_2019_orig_2}. This reduces ingredients to more basic forms; for instance, “finely grated cheese” and “coarsely grated cheese” both become “grated cheese” with this merging scheme. After this initial ingredient preprocessing, we were left with a set of around 1,500 ingredients.

However, after an initial model evaluation, we realized that further ingredient refinement was required to improve model predictions. To accomplish this, certain word fragments and filler words (\eg "and," "or," and "the") were removed. Additionally, adjectives of the ingredients themselves were completely removed to the best of our ability. For example, words like "creamy", "superfine", and "freshly" did not add value to their base ingredients and instead only added complexity for our model to produce useful ingredient predictions. Cooking utensils and containers like "jars" and "skillet" are generally useful when preparing a recipe, but we chose to remove these as ingredients and instead focused on edible ingredients.

We found that adjectives involving the color, origin, and quantity of ingredients also contributed to the complexity of ingredient prediction. A decision was made that color remains an important distinction, like that of red peppers versus green peppers, but geographic adjectives (such as country of origin) were of lesser importance. We combine occurrences of quantity (\eg "1 pound bananas" or "2 cups bananas") and plurals (\eg "carrot" versus "carrots") into single ingredients. Finally, we further restrict the list of ingredients to the top 1\% frequently occurring ingredients. After these refinements, our ingredients list consisted of around 200 mostly unique items.

\subsubsection{Ingredient model architecture}

For ingredient prediction, we compared a custom CNN against a model that utilized a pre-trained ResNet-50 model as a feature extractor. We were inspired to do this through lectures in class as well as the research performed by researchers in \cite{recipe1707}. Here, researchers used a ResNet-50 model that was pre-trained on ImageNet and showed improved performance in metrics. Rather than multi-class classification, our ingredient prediction is framed as a multi-label classification problem where the outputs of each model represent confidences for each ingredient. Both models receive a 200-by-200 food image as input.

Our custom CNN model architecture utilizes multiple convolutional blocks followed by a flatten layer, a fully-connected hidden layer with 256 neurons, and a fully-connected ingredient prediction layer. Each convolutional block is composed of the following sequence of layers: a 3x3 convolution with an increasing number of filters in each block and ReLU activation, a batch normalization layer, and a 2x2 max pooling layer (See Figures 10 and 11 in the appendix). This model also utilized L2 regularization and dropout as a measure to boost generalization and alleviate overfitting.

Our ResNet-50-based model architecture takes advantage of transfer learning by reusing ResNet-50 trained on the ImageNet dataset. The ResNet-50 model is frozen and it's classification layer is removed and replaced with a dropout layer and a trainable fully-connected ingredient prediction layer.

The custom CNN architecture has learned parameters in its convolutional, fully connected, and batch normalization layers. Its pooling and flatten layers do not contain any learned parameters. Within the ResNet-50 architecture, the final ingredient prediction layer is the only layer that contains learnable parameters, as we utilized ResNet-50 as a fixed feature extractor. This means that despite our custom model having fewer parameters overall, it had more learnable parameters than our ResNet-50-powered model.

As both of our models were tasked with predicting ingredients from images, we employ sigmoid activation in the output layers of each model and binary cross-entropy as our loss function, as it is suitable for multi-label classification scenarios. Additionally, both of our models were updated using the Adam optimizer, as it tends to converge faster \cite{kingma2017adam}.

\subsubsection{Ingredient model training procedure}

Multiple model architectures and hyperparameters were trained and evaluated for both our custom CNN model and our ResNet-50-powered model. While experimenting, accuracy proved to be a poor evaluation metric due to an imbalance in ingredient occurrence. Due to this, F1 score was chosen as one of our evaluation metrics. As in \cite{Salvador_Drozdzal_Giro-i-Nieto_Romero_2019_orig_2}, we also employed Intersection over Union (IoU) as an evaluation metric, as it is an intuitive measure of multi-label classification performance. During experimentation, we noticed the shapes of both F1 and IoU curves were similar, and so we ultimately decided on IoU as our primary ingredient prediction metric.

When accuracy is low and F1 and IoU scores are comparatively better, it suggests that our model might be handling certain ingredients well while struggling with others. This scenario is common in imbalanced datasets or when some classes are inherently more challenging for the model to predict accurately. F1 score and IoU, being more sensitive to minority classes, can better reflect the model's performance. F1 score and IoU are also threshold-dependent metrics, meaning that they can vary based on the threshold used to determine positive predictions. In our usage of these metrics, we chose 0.5 (a 50\% confidence rate) as our threshold.

Our dataset was first split into a training and testing set at a 80/20 ratio. 20\% of the training set was further reserved as validation data. Model hyperparameters were chosen uniformally using random search, and validation IoU scores were compared across all runs. Each model was trained for a maximum of 25 epochs on the training set, and early stopping was employed to prevent overfitting.

Following suggestions left by researchers in \cite{recipe1707} and \cite{Salvador_Drozdzal_Giro-i-Nieto_Romero_2019_orig_2}, as well as utilizing methods taught in our lectures, we were fairly confident in our approach. However, throughout our experiments, we realized that data cleaning and hyperparameter tuning would have a large impact on our model's performance.





\subsection{Stage 2: Instruction Generation}

\subsection{Model architecture}
In Stage 2, our goal was to train a Long-Short Term Memory (LSTM) model to provide coherent and contextually relevant cooking instructions based on an input ingredient list, serving as a valuable tool for recipe generation. While several instruction generation architectures were explored, we primarily compared the performance of two models. Our first instruction generation model learns a set of embeddings and uses a standard LSTM layer. Our second model utilizes pre-trained GloVe embeddings \cite{pennington-etal-2014-glove}, a bidirectional LSTM layer, and also employs L2 regularization, dropout, and layer normalization. Both models include a final time distributed fully-connected layer for instruction token prediction.

\subsubsection{Data Processing}
Two tokenizers are created: one for the ingredient vocabulary, and one for the instruction vocabulary. Sequences of ingredients and instructions are then transformed by their respective tokenizers, and are padded to the length of the longest sequence in each dataset, ensuring a uniform sequence length. A custom data generator is implemented for efficient batch processing during training.

\subsubsection{Instruction Generator Training Details}
The model is trained using the Adam optimizer, with categorical cross-entropy used as the loss function. In addition to accuracy, perplexity was chosen as a validation metric 
\cite{10.1121/1.2016299_orig_4}. Both metrics are monitored throughout training and early stopping is employed to halt training if validation perplexity does not improve after three consecutive epochs, with the best weights restored.

\section{Experiments and Results}

\subsection{Stage 1 --- Ingredient Prediction}
Ingredient prediction model hyperparameters were explored using random search. Our goal in our searches were to find a set of parameters for which we could achieve at least 0.1 validation IoU. Although this seems low, we realized that this would still be difficult for us. For both ResNet-50 and custom architectures, 30 individual training runs were performed. After each training run, metrics and model weights were stored for evaluation and comparison.

\subsubsection{Custom CNN Architecture}
For our custom CNN architecture, we explored batch sizes of 32, 64, and 128, the number of convolutional blocks used (3, 4, or 5), learning rates of 1e-3, 1e-4, and 1e-5, image augmentation, and regularization. For this architecture, the hyperparameter "regularization" referred to the presence of both a dropout rate of 0.7 before the final layer and L2 regularization of 1e-3 on the hidden fully-connected layer. When image augmentation was disabled, only image rescaling was applied to ensure input image sizes were uniform. Mean per-hyperparameter-value IoU scores can be seen in Figure 1.

When reviewing results for our custom CNN architecture, we find that a lower batch size generally produced a higher validation IoU score. This could be due to the additional training steps per epoch offered by a lower batch size, or it could be that the natural regularizing effect of a lower batch size improved generalization of the model. We also find that a larger number of convolutional blocks greatly improves performance. This is likely due to the more complex filter representations offered by deeper convolutional models. Finally, we find that our explicit regularization scheme hindered our model's validation accuracy. It is likely that our regularization amounts (an L2 regularization rate of 1e-3 and a dropout rate of 0.7) limited our weights and hindered information transfer to the final classification layer.

\begin{figure}[t]
\begin{center}
   \includegraphics[width=0.8\linewidth]{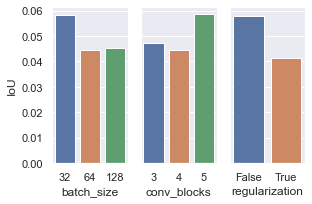}
\end{center}
\caption{Average IoU scores for our custom CNN architecture by select hyperparameter values}
\label{fig:onecol}
\end{figure}

After this search completed, the optimal set of hyperparameters was determined to be a batch size of 128, 4 convolutional blocks, a learning rate of 1e-3, with both augmentation and regularization disabled. This configuration resulted in an validation IoU score of 0.075. However, this custom CNN model was not successful in reaching our goal of ingredient prediction.

\begin{figure}[t]
\begin{center}
   \includegraphics[width=0.7\linewidth]{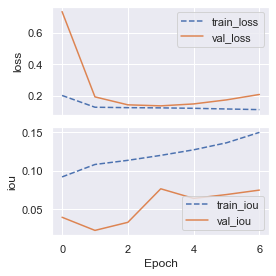}
\end{center}
\caption{Training and validation metrics for the best found hyperparameters set on our custom CNN architecture, by epoch}
\label{fig:onecol}
\end{figure}

\subsubsection{ResNet-50 Architecture}
For our ResNet-50-powered architecture, we searched batch sizes of 32, 128, and 512, learning rates of 1e-3, 1e-4, and 1e-5, augmentation, and dropout rates of 0, 0.3, and 0.7. Mean IoU scores for each hyperparameter value IoU scores can be seen in Figure 3.

We find that batch size had a lesser effect on this architecture than with our custom CNN architecture, although a lower batch size of 32 still produced a higher validation IoU on average. Learning rate, however, proved to be a very influential hyperparameter, resulting in an increase to validation IoU by around 50\% on average between the lowest and highest learning rates tested. A larger learning rate results in larger gradient steps taken during weight updates. A combination of the Adam optimizer and utilizing ResNet-50 as a fixed feature extractor may have allowed the final ingredient prediction layer to take these larger gradient steps while still settling in an acceptable minimum. Finally, image augmentation did not play as large of a role as we had hoped across either model architecture. This could be due to our task being a multi-label classification problem. Another potential explanation is that our training data was adequate varied when compared to our validation data.

\begin{figure}[t]
\begin{center}
   \includegraphics[width=0.8\linewidth]{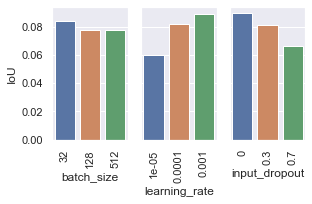}
\end{center}
\caption{Average IoU scores for our ResNet-50 architecture by select hyperparameter values}
\label{fig:onecol}
\end{figure}

The hyperparameter search for the ResNet-50-powered architecture resulted in a batch size of 512, dropout of 0, and learning rate of 1e-3. This hyperparameter set resulted in a validation IoU score of 0.106. This CNN achieved the goal set by our team. It is important to note that although validation scores may be higher for some hyperparameter values, it does not guarantee that the combination of all the highest-scoring values will result in a superior model. Comparisons such as these therefore offer more insight when analyzing performance across a single hyperparameter's values.

\begin{figure}[t]
\begin{center}
   \includegraphics[width=0.7\linewidth]{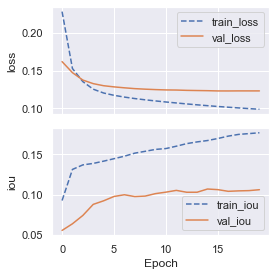}
\end{center}
\caption{Training and validation metrics for the best found hyperparameters set on our ResNet-50-powered architecture, by epoch}
\label{fig:onecol}
\end{figure}

\subsubsection{Comparing Ingredient Prediction Architectures}
As expected, utilizing a pre-trained ResNet-50 model provided higher performance than our custom CNN model, given the amount of training time and resources available. For this reason, we selected the ResNet-50 model for final ingredient predictor evaluation purposes. It is possible that a custom model architecture could outperform our transfer learning approach. However, the ability to reuse existing models to reduce training time can not be understated.

On the test dataset, our ResNet-50 model obtained an IoU score of 0.119, which is higher than its performance on the validation set.

We use a comparatively low confidence threshold of 0.05 when performing ingredient prediction, as when using the validation threshold of 0.5, too few ingredients are predicted for each input image. When analyzing ingredient prediction results, we find that our model performs best when identifying presence of garnishes. However, our model's performance suffers from over-predicting common ingredients such as salt, sugar, olive oil, and lemon juice. One notable result, however, is that our model identified that a recipe contains water, although the true recipe ingredients did not mention water (See Table 2 in appendix).

\subsection{Stage 2 --- Instruction Generation}
In stage 2, our focus is on generating a coherent set of instructions that can be used to recreate a recipe with the given ingredients. The evaluation of our generated recipe steps involves both human judgment and automated metrics. The primary automated metric we chose to employ for our instruction generation task is perplexity. Perplexity is calculated through cross-entropy and serves as an automated measure of a model's predictive capability. It quantifies the average uncertainty of the model for each dataset sample, with lower perplexity indicating superior predictive performance. Our goal was to achieve a validation perplexity of below 500.

In the absence of a pre-trained model, a significant gap between training and validation perplexity would suggest overfitting. To mitigate this, we utilize pre-trained GloVe embeddings, observing a decrease in both training and validation loss perplexity with increasing epochs. Results of this can be seen in Figures 5 and 6.

\subsubsection{Model Architecture Experiments}

We experiment with a varying number of LSTM units, pre-trained GloVe embeddings, learning rate, batch size, and several regularization methods. We explored LSTM sizes of 8, 16, 32, and 64 units, learning rates of 1e-1, 1e-3, 5e-4, and 1e-4, and batch sizes of 8, 16, 32, and 64. For regularization methods, we explored the effects of dropout, L2 regularization, and layer normalization.

Through our experimentation, we find GloVe embedding sizes of both 50 and 100 result in better performance over a model without pre-trained GloVe embeddings. There was no noticeable difference in validation metrics between an embedding size of 50 or 100, so 50 was chosen in favor of model execution speed. This performance increase is expected, as GloVe embeddings map words into a global space by a semantic similarity metric.

While experimenting with our LSTM layer, we find that a bidirectional LSTM offers better performance than a standard LSTM layer. This could be due to our treating ingredients as a set, rather than as a list. By using a bidirectional layer, more context from both ends of the ingredient sequence is preserved. We also find through experimentation that an LSTM size of 8 units was optimal for our task; higher values led to increased memory usage and higher perplexity scores.

We find that by employing a dropout rate of 0.8, L2 regularization of 1e-2, and layer normalization, our validation perplexity can be further improved. These regularization methods prevent the LSTM model from overfitting to our training data and offer better generalization capabilities.

\subsubsection{Model comparison}

Our LSTM model that did not use GloVe embeddings repeatedly generated common words such as "and" and "the". Comparatively, we find that by employing GloVe embeddings and regularization methods, a much better validation perplexity and generative output can be achieved. Results of this can be found in Table 1 in the Appendix. For this second model, the optimal set of hyperparameters was a learning rate of 1e-2, a batch size of 64, and a training length of 15 epochs. These improvements resulted in a validation perplexity of 434.413 --- an increase over our model without pre-trained GloVe embeddings, which had a validation perplexity of 529.227. The model that used pre-trained embeddings achieved our goal of under 500 validation perplexity while the model without pre-trained embeddings did not.

\begin{figure}[t]
\begin{center}
   \includegraphics[width=1\linewidth]{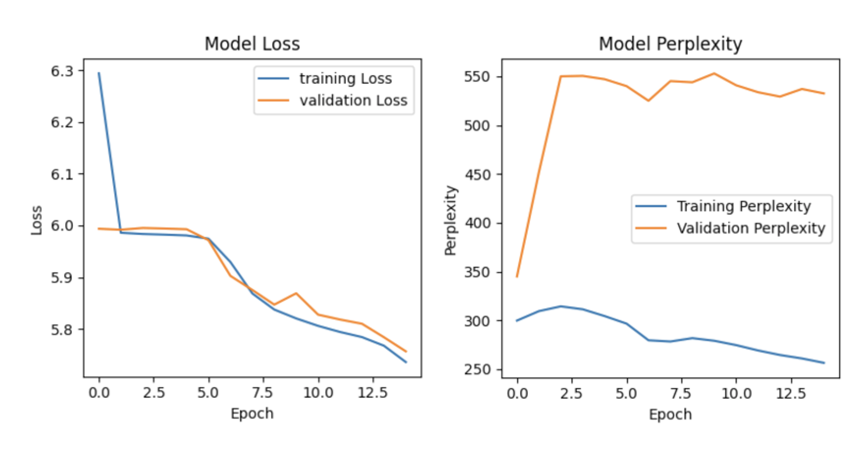}
\end{center}
   \caption{LSTM without Using Pre-trained Model}
\label{fig:onecol}
\end{figure}

\begin{figure}[t]
\begin{center}
   \includegraphics[width=1\linewidth]{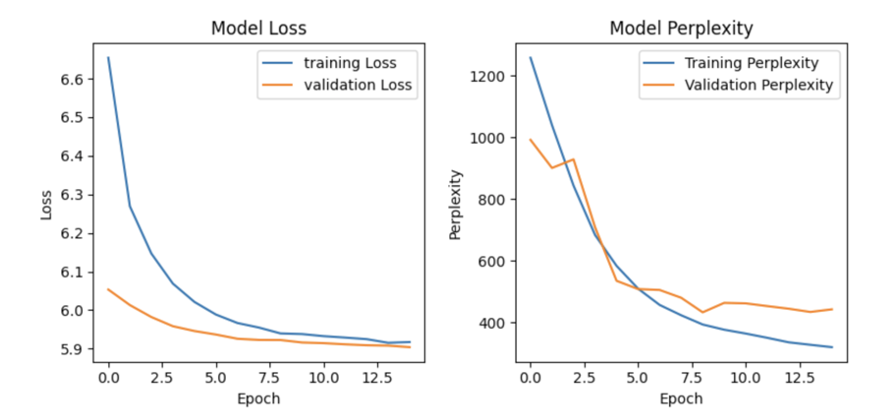}
\end{center}
   \caption{LSTM Using Pre-trained Model}
\label{fig:onecol}
\end{figure}

\section{Conclusion}

\subsection{Dataset Considerations}

For our LSTM model, the small gap between training and validation perplexity indicates no overfitting. However, our limited dataset size (approximately 240 Megabytes) is acknowledged. While our pragmatic compromise considers hardware limitations, utilizing a larger training dataset often benefits deep learning models and should produce better results on both ingredient prediction and instruction generation tasks.

One issue that heavily impacted ingredient prediction was that of class imbalance. By taking additional measures to mitigate the effects of class imbalance, a more robust Image-to-Recipe Translation model could be produced. One measure to investigate is class weighting, which decreases the impact of common classes on the loss value, and increases the impact of rarer classes.

\subsection{Future Steps and Model Consideration}

In addition to an increased dataset size, one future initiative for instruction generation is to explore advanced architectures such as transformer-based models. The study "Inverse Cooking: Recipe Generation from Food Image" \cite{Salvador_Drozdzal_Giro-i-Nieto_Romero_2019_orig_2} underscores their efficacy on larger datasets. In particular, by utilizing an R Transformer-based architecture, a text instruction generation model could be trained on datasets of a much larger scale.

Additional future work includes model deployment and user feedback. Once a performant Image-to-Recipe Translation model is developed, it could be deployed as a web or mobile application for others across the world to use. Users could then provide valuable feedback or even contribute data to further improve the model.

In conclusion, transfer learning and dataset size are very influential in the task of Image-to-Recipe Translation. Architectures, such as transformers, represent a viable path for further improving instruction generation. These factors are likely to play significant roles in determining the evolution of recipe instruction generation models as the fields of CV and NLG progress.

{\small
\bibliographystyle{ieee_fullname}
\bibliography{egbib}
}

\clearpage  

\appendix
\section{APPENDIX}


\begin{figure}[ht]
\begin{center}
   \includegraphics[width=0.8\linewidth]{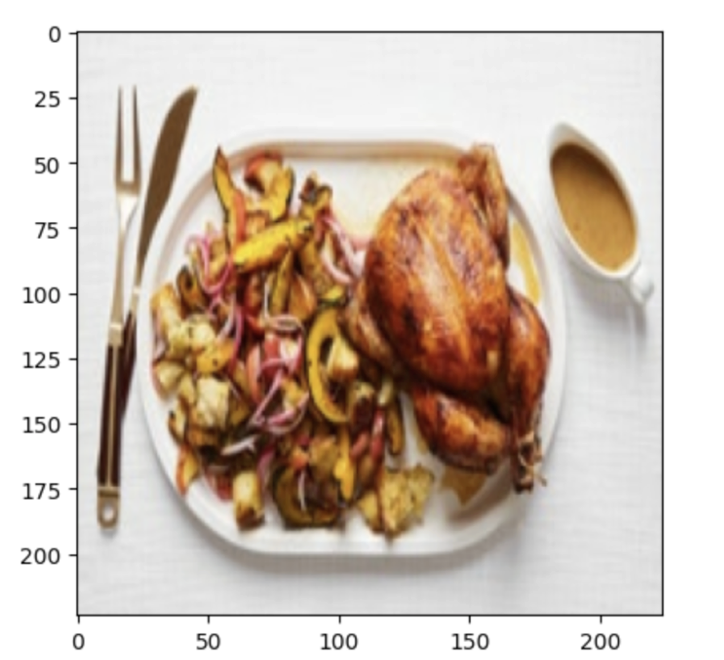}
\end{center}
\caption{Stage 2 Tested Food Image}
\label{fig:onecol}
\end{figure}


\begin{table*}
    \centering
    \begin{tabular}{|p{2cm}|c|p{5cm}|p{7cm}|}
        \hline
        Model & Perplexity  & Input Ingredients & Predicted Output \\
        \hline\hline
        LSTM without Using Pre-trained Model & 532.5060 & chicken,
        salt,
        acorn squash,
        squash,
        sage,
        rosemary,
        butter,
        salt,
        allspice,
        pepper,
        black pepper,
        pepper,
        bread,
        pie,
        white bread,
        apple,
        pie,
        gin,
        oil,
        olive,
        olive oil,
        onion,
        red onion,
        apple,
        cider,
        vinegar,
        miso,
        flour,
        butter,
        salt,
        white wine,
        wine,
        broth,
        chicken,
        salt,
        miso,
        salt,
        pepper & a n d   a n d   a n d   a n d   a n d   a n d   a n d   a n d   a n d   a n d   a n d   a n d   a n d   a n d   a n d   a n d   a n d   a n d   a n d   a n d   a n d   a n d   a n d   a n d   a n d   a n d   a n d   a n d   a n d   a n d a n d   a n d   a n d   a n d   a n d   a n d   a n d   a n d   a n d   a n d   a n d   a n d   a n d   a n d   a n d   a n d   a n d   a n d   a n d   a n d   a n d   a n d   a n d   a n d   a n d   a n d   a n d   a n d   a n d   a n d   a n d   a n d   a n d   a n d  \\
        LSTM Using Pre-trained Model & 442.9162 & chicken,
        salt,
        acorn squash,
        squash,
        sage,
        rosemary,
        butter,
        salt,
        allspice,
        pepper,
        black pepper,
        pepper,
        bread,
        pie,
        white bread,
        apple,
        pie,
        gin,
        oil,
        olive,
        olive oil,
        onion,
        red onion,
        apple,
        cider,
        vinegar,
        miso,
        flour,
        butter,
        salt,
        white wine,
        wine,
        broth,
        chicken,
        salt,
        miso,
        salt,
        pepper & and the to a in with until 1 minutes of 2 add heat over about bowl salt into on medium then 4 large oil or 3 for is cook water mixture transfer pan pepper oven at remaining from baking stir let sugar place cup cover butter stirring cool top it skillet 5 if small remove be season using dough each ahead inch whisk high serve are set sheet sauce cut brown up preheat can side garlic 10 pot bring occasionally golden boil toss combine saucepan flour sprinkle bake just out simmer chill cream rack pour juice temperature spoon chicken tablespoons through as hours all more lemon tender half together you teaspoon mix 8 smooth grill an sides lightly 30 room heavy do will paper low 6 not cake drain beat onion hot 15 made egg cups before coat liquid slightly well tablespoon dry bottom gently reduce warm off center but down cheese spread minute aside least them stand by ice ingredients your 20 dish pieces eggs turn surface keep cooking foil parchment onto wrap thick divide very arrange vinegar cooked hour blend serving when milk slices chocolate 12 use drizzle potatoes any browned brush meanwhile plate make seeds completely evenly taste turning covered one put pasta layer time bread tbsp roast rimmed meat another processor discard sauté syrup roll platter knife some filling batter form powder peel leaves tomatoes cold slice vanilla rice around fold once broth skin olive return fish onions vegetables soft pork reserved crisp prepared edges  \\
        \hline
    \end{tabular}
    \caption{Comparison of stage 2 models' output (See Figure 7)}
    \label{tab:contributions}
\end{table*}

\begin{figure}[ht]
\begin{center}
   \includegraphics[width=0.8\linewidth]{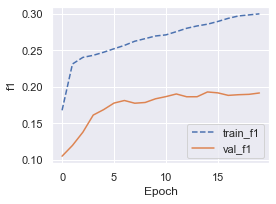}
\end{center}
\caption{ResNet-50 model F1 scores}
\label{fig:onecol}
\end{figure}

\begin{figure}[ht]
\begin{center}
   \includegraphics[width=0.8\linewidth]{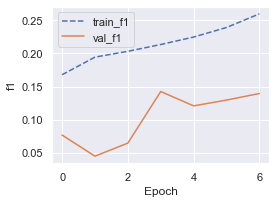}
\end{center}
\caption{Our custom CNN model F1 scores}
\label{fig:onecol}
\end{figure}

\begin{table*}
    \centering
    \begin{tabular}{|p{5cm}|p{5cm}|p{5cm}|}
        \hline
        Image & True Ingredients & Predicted Ingredients \\
        \hline\hline
        \includegraphics[width=1\linewidth]{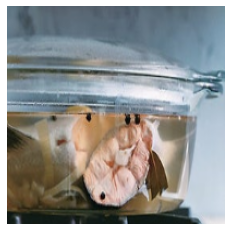} & 
        allspice, bay leaf, black peppercorns, dijon mustard, dill, lemon zest, mayonnaise, onion, sour cream, tarragon &
        black pepper, garlic cloves, lemon juice, salt, sugar, unsalted butter, water, whipping cream \\
        \hline
        \includegraphics[width=1\linewidth]{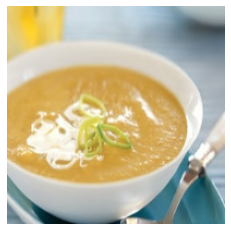} & 
        coriander seeds, creme fraiche, lemon peel, olive oil, sour cream, vegetable broth &
        black pepper, garlic cloves, heavy cream, lemon juice, lemon zest, lime juice, olive oil, parsley, salt, sugar \\
        \hline
        \includegraphics[width=1\linewidth]{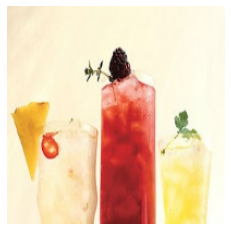} & 
        ice, lime juice, simple syrup &
        black pepper, lemon juice, lime juice, milk, olive oil, salt, sugar \\
        \hline
    \end{tabular}
    \caption{Select ingredient predictions and true ingredient list}
    \label{tab:contributions}
\end{table*}

\begin{figure}[ht]
\begin{center}
   \includegraphics[width=0.8\linewidth]{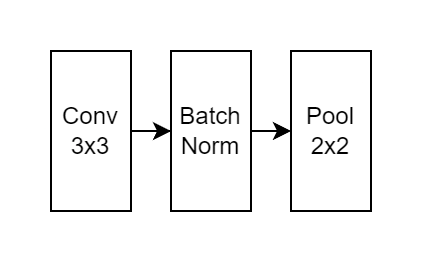}
\end{center}
\caption{Our custom CNN model's convolutional block consisting of 3x3 conv, batch norm, and 2x2 max pooling layers}
\label{fig:onecol}
\end{figure}

\begin{figure}[ht]
\begin{center}
   \includegraphics[width=1\linewidth]{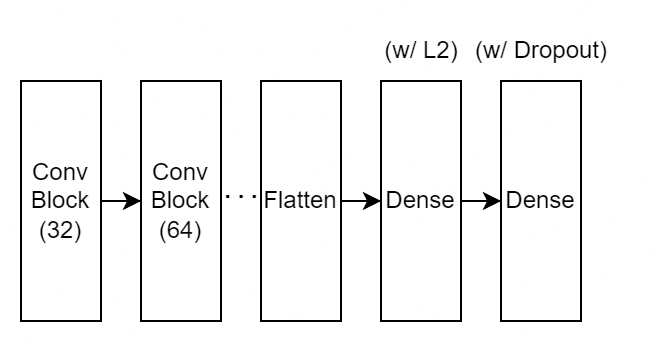}
\end{center}
\caption{An example configuration of our custom CNN model architecture. Each model tested consisted of 3-5 convolutional blocks, followed by flatten and two fully-connected layers.}
\label{fig:onecol}
\end{figure}

\begin{figure}[ht]
\begin{center}
   \includegraphics[width=1\linewidth]{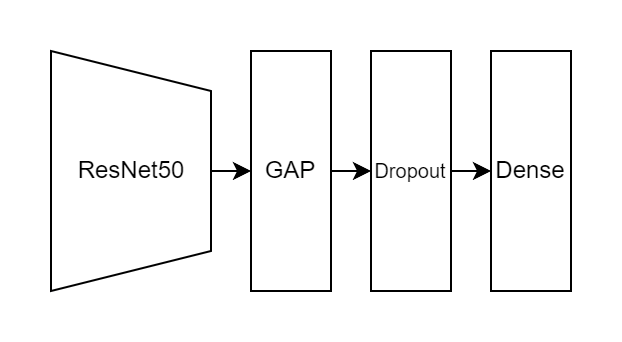}
\end{center}
\caption{Our ResNet-50 model architecture using ResNet-50 as a fixed feature extractor followed by global average pooling and a final fully-connected ingredient prediction layer}
\label{fig:onecol}
\end{figure}


\end{document}